\documentclass[conference]{IEEEtran}
\IEEEoverridecommandlockouts
\usepackage{subfigure} 
\usepackage{cite}
\usepackage{amsmath,amssymb,amsfonts}
\usepackage{algorithmic}
\usepackage{graphicx}
\usepackage{textcomp}
\usepackage{xcolor}
\usepackage{mathrsfs}

\def\BibTeX{{\rm B\kern-.05em{\sc i\kern-.025em b}\kern-.08em
    T\kern-.1667em\lower.7ex\hbox{E}\kern-.125emX}}
\begin{document}

\title{Real-Time Stereo Vision for Road\\ Surface 3-D Reconstruction}

\author{Rui Fan$^{1,2}$, Yanan Liu$^{3}$, Xingrui Yang$^{2}$, Mohammud Junaid Bocus$^{2}$, Naim Dahnoun$^{4}$, Scott Tancock$^{4}$\\
$^{1}$Robotics Institute, The Hong Kong University of Science and Technology, Hong Kong SAR, China.\\
$^{2}$Visual Information Institute, University of Bristol, Bristol, BS8 1UB, United Kingdom.\\
$^{3}$Bristol Robotics Laboratory, University of Bristol, Bristol, BS16 1QY, United Kingdom.\\
$^{4}$Department of Electrical and Electronic Engineering, University of Bristol, Bristol, BS8 1UB, United Kingdom.
}

\maketitle

\begin{abstract}

Stereo vision techniques have been widely used in civil engineering to acquire 3-D road data. The two important factors of stereo vision are accuracy and speed. However, it is very challenging to achieve both of them simultaneously and therefore the main aim of developing a stereo vision system is to improve the trade-off between these two factors. In this paper, we present a real-time stereo vision system used for road surface 3-D reconstruction. The proposed system is developed from our previously published 3-D reconstruction algorithm where the perspective view of the target image is first transformed into the reference view, which not only increases the disparity accuracy but also improves the processing speed. Then, the correlation cost between each pair of blocks is computed and stored in two 3-D cost volumes. To adaptively aggregate the matching costs from neighbourhood systems, bilateral filtering is performed on the cost volumes. This greatly reduces the ambiguities during stereo matching and further improves the precision of the estimated disparities. 
Finally, the subpixel resolution is achieved by conducting a parabola interpolation and the subpixel disparity map is used to reconstruct the 3-D road surface. The proposed algorithm is implemented on an NVIDIA GTX 1080 GPU for the real-time purpose. The experimental results illustrate that the reconstruction accuracy is around 3 mm.
\end{abstract}

\begin{IEEEkeywords}
stereo vision, 3-D reconstruction, bilateral filtering, subpixel disparity map, real-time.
\end{IEEEkeywords}

\section{Introduction}
\label{sec.introduction}
\IEEEPARstart{T}HE condition assessment of asphalt pavements is essential to ensure their serviceability while still providing maximum traffic safety \cite{Koch2015}. However, the manual visual inspections performed by either structural engineers or certified inspectors are hazardous, time-consuming and cumbersome \cite{Kim2014}. 
Over the past decade, various technologies such as active sensing and passive sensing have been increasingly utilised in civil engineering to reconstruct the 3-D road surface and assist the personnel in assessing the physical and functional condition of asphalt infrastructures.  
These 3-D reconstruction methods can mainly be classified as laser scanner-based \cite{Tsai2017}, Microsoft Kinect-based \cite{Jahanshahi2012} and passive sensor-based \cite{Fan2018}. Laser scanning equipment records the  laser pulses reflected from an object to acquire its accurate  3-D model \cite{Schnebele2015}. However, such laser scanners are not widely used because both the equipment and its long-term maintenance are very expensive \cite{Koch2011}. Furthermore, the Microsoft Kinect sensors greatly suffer from the infrared saturation in direct sunlight and materials which can strongly absorb the infrared light also make the depth information unreliable \cite{Cruz2012a}. Therefore, the Microsoft Kinect-based methods are somewhat  ineffective
in outdoor environments while the passive sensors are more capable of reconstructing the road surface for condition assessment \cite{Sadjadi2010, Fan2016}.

To reconstruct a real-world environment using passive sensors, multiple camera views are required \cite{Hartley2003, Ma2018}. Images from different viewpoints can be captured using either a single moveable camera or a group of synchronised cameras \cite{Sadjadi2010}. In this paper, we use a ZED camera \cite{ZED} to acquire stereo image pairs for 3-D reconstruction. Since the stereo rig is assumed to be well-calibrated, the main work performed in this paper is disparity estimation. 

The two most important performance factors in disparity estimation are speed and accuracy \cite{Fan2017}. A lot of research has been carried out over the past decade to improve both the precision of the disparity maps and the execution speed of the algorithms \cite{Tippetts2012}. {However,  the disparity estimation algorithms designed to achieve better disparity accuracy usually have higher computational complexity and lower processing efficiency \cite{Fan2017}}. Hence, speed and accuracy are two desirable but conflicting properties, and it is very challenging to achieve both of them simultaneously \cite{Tippetts2016}. Therefore, the main motivation of developing a disparity estimation algorithm is to improve the trade-off between accuracy and speed \cite{Tippetts2016}.

\begin{figure*}[!t]
	\begin{center}
		\centering
		\includegraphics[width=0.95\textwidth]{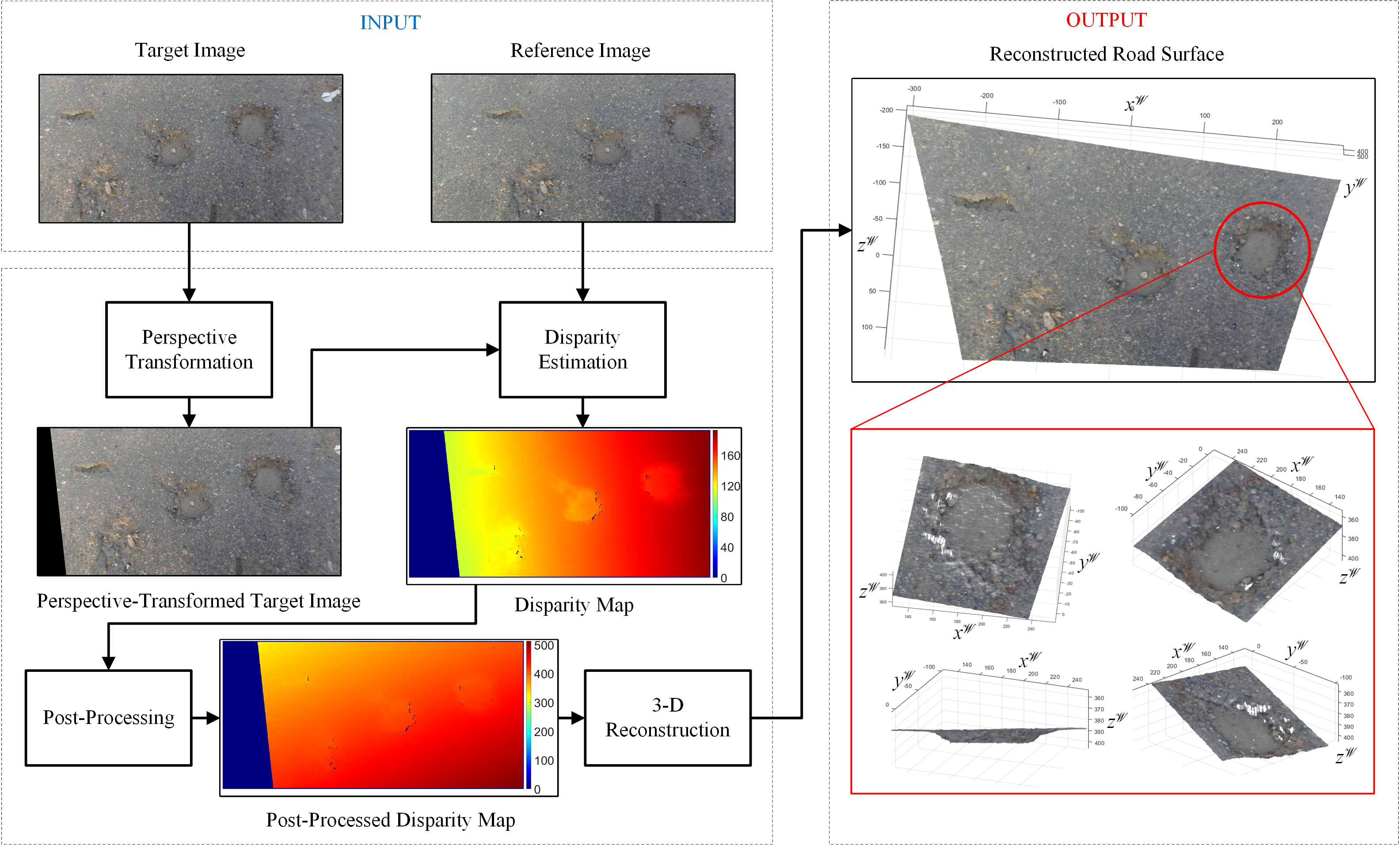}
		\caption{Overview of the proposed road surface 3-D reconstruction system.}
		\label{fig.block_diagram}
	\end{center}
\end{figure*}

The state-of-the-art algorithms for disparity estimation are mainly classified as local, global and semi-global \cite{Fan2018}.  The local algorithms simply match a series of blocks from the left and right images and select the correspondence with the lowest cost or the highest correlation. This optimisation is also known as Winner-Take-All (WTA) \cite{Fan2017}. Unlike the local algorithms, global algorithms process the stereo matching using some sophisticated techniques such as Belief Propagation (BP) \cite{Ihler2005} and Graph Cut (GC) \cite{Boykov2001}. These techniques are usually developed based on Markov Random Fields (MRF), where the process of finding the best disparities translates to a probability maximisation problem \cite{Szeliski2008, Tappen2003, Blake2011}. Semi-global matching (SGM) approximates the MRF inference by performing cost aggregation along all directions in the image and this greatly improves the precision and efficiency of stereo matching \cite{Hirschmuller2008, Spangenberg2013}. However, finding the optimum values for the smoothness parameters is a difficult task due to the occlusion problem. Over-penalising the smoothness term can reduce the ambiguities around the discontinuities but on the other hand can cause errors for continuous areas \cite{Fan2018}. In \cite{Mozerov2015}, Mozerov and Weijer proved that the initial energy optimisation problem in a fully connected MRF model can be formulated as an adaptive cost aggregation using bilateral filtering. Since then, a lot of endeavours have been made in local algorithms \cite{Hosni2013} to improve the accuracy of the disparity maps by performing bilateral filtering on the cost volume before estimating the disparities. These algorithms are also known as fast bilateral stereo (FBS). However, filtering the whole cost volume is always computationally intensive  and therefore the FBS must be implemented on powerful hardware for the real-time purpose  \cite{Fan2018}. 

This paper presents a real-time road surface 3-D reconstruction system which is developed based on fast bilateral stereo. The block diagram of the proposed system is illustrated in Fig. \ref{fig.block_diagram}. First, the perspective view of the target image is transformed into the reference view, which greatly enhances the similarity of the road surface between the reference and target images. In this paper, the left and right images are considered as the reference and target images, respectively. Then, Normalised Cross-Correlation (NCC) is used to measure the similarity between each pair of selected blocks. The computed correlation costs are stored in two 3-D cost volumes. To adaptively aggregate the correlation costs from the neighbourhood systems, we perform bilateral filtering on the two cost volumes. This significantly reduces the ambiguities during stereo matching and enhances the accuracy of the estimated disparities. Finally, the estimated disparity map is post-processed and the 3-D road surface is reconstructed.

The remainder of this paper is organised as follows: Section \ref{sec. algorithm_description} describes the proposed road surface 3-D reconstruction algorithm. In Section \ref{sec.experimental_results}, the experimental results are illustrated and the evaluation of the proposed system is carried out. Finally, Section \ref{sec.conclusion_future_work} summaries the paper and provides some recommendations for future work.

\section{Algorithm Description}
\label{sec. algorithm_description}

\subsection{Perspective Transformation}
\label{sec.perspective_transformation}

In \cite{Fan2018}, we treated the road surface as a ground plane. For an arbitrary 3-D point $\boldsymbol{p^\mathscr{W}}=[x^\mathscr{W},y^\mathscr{W}, z^\mathscr{W}]^\top$ on the road surface, its projections $\boldsymbol{p_l}=[u_l,v_l]^\top$ on the left image and $\boldsymbol{p_r}=[u_r,v_r]^\top$ on the right image can be linked using a so-called homograph matrix $\boldsymbol{H}$ as follows \cite{Hartley2003}:

\begin{equation}
\boldsymbol{\tilde{p}_r}=\boldsymbol{H}\boldsymbol{\tilde{p}_l}
\label{eq.homograph_mat}
\end{equation}
where $\boldsymbol{\tilde{p}}=[\boldsymbol{p}^\top,1]^\top$ represents the homogeneous coordinate of $\boldsymbol{p}$. $\boldsymbol{H}$ can be estimated with at least four pairs of correspondences $\boldsymbol{p_l}$ and $\boldsymbol{p_r}$ \cite{Hartley2003}. 

For a well-rectified stereo image pair, $v_l=v_r=v$ and the disparity $d$ is defined as $d=u_l-u_r$. With the assumption that the road surface is a horizontal plane, i.e., $n y^\mathscr{W}+\beta=0$, the relationship between each pair of  correspondences $\boldsymbol{p_l}$ and $\boldsymbol{p_r}$ can be represented as a linear model as follows \cite{Fan2018}:

\begin{equation}
d=(v_0\cos\theta-f\sin\theta)\frac{T_cn}{\beta}-\cos\theta\frac{T_cn}{\beta}v=\alpha_0+\alpha_1 v
\label{eq.relation_vd}
\end{equation}
where $f$ is the focal length of the left and right cameras, $\theta$ is the angle between stereo rig and road surface, $T_c$ is the baseline, and $(u_0, v_0)$ is the principal point in pixels.

Now, the perspective transformation can be straightforwardly realised using a parameter vector $\boldsymbol{\alpha}=[\alpha_0, \alpha_1]^\top$. The latter can be estimated by solving a least squares problem with a collection of reliable correspondence pairs $\boldsymbol{p_l}$ and $\boldsymbol{p_r}$. It is to be noted that the roll angle $\gamma$ for the initial frame needs to estimated to minimise its impact on the perspective transformation for the rest of the sequences. As in \cite{Ozgunalp2017}, $\gamma$ can be estimated by fitting a linear plane $d(u,v)=\gamma_0+\gamma_1u+\gamma_2v$ to a small patch from the near field in the disparity map and $\gamma=\arctan(-\gamma_1/\gamma_2)$. Then, we utilise ORB (Oriented FAST and Rotated BRIEF) \cite{Rublee2011} to detect and match the  correspondence pairs. Since outliers can severely affect the accuracy of least squares fitting (LSF), we apply Random Sample Consensus (RANSAC) in the LSF to minimise the effects caused by the outliers. More details on $\boldsymbol{\alpha}$ estimation are provided in \cite{Fan2018a}.  
Then, each point on row $v$ in the target image is shifted $\alpha_0+\alpha_1v-\delta$ pixels to the right, where $\delta$ is a constant set to guarantee that all the disparities are non-negative. An example of the perspective-transformed target image is illustrated in Fig. \ref{fig.block_diagram}. 

\subsection{Disparity Estimation}
\label{sec.disp_estimation}

A disparity estimation algorithm usually consists of four steps: cost computation, cost aggregation, disparity optimisation and disparity refinement \cite{Mozerov2015}. { In the following subsections, we elaborate on each step of the proposed disparity estimation algorithm.}

\subsubsection{Cost Computation}

In stereo matching, the most commonly used approaches for cost computation are Sum of Absolute Difference (SAD) and Sum of Squared Difference (SSD) \cite{Scharstein2011}. However, these two approaches are very sensitive to the intensity difference between the left and right images, which may further lead to some incorrect matches in the process of disparity estimation. In this paper, we use  NCC as the cost function to measure the similarity between each pair of blocks selected from the left and right images. Although  NCC is more computationally intensive than  SAD and SSD, it can  provide more accurate results when an intensity difference is involved \cite{Fan2018a}. The cost function of  NCC is as follows \cite{Fan2017}:
\begin{equation}
c(u,v,d)=\frac{\sum\limits_{x=u-\varrho}^{x=u+\varrho}\sum\limits_{y=v-\varrho}^{y=v+\varrho} i_{l}(x,y) i_{r}(x-d,y)-n\mu_{l} \mu_{r}}{n\sigma_l \sigma_r}
\label{eq.ncc}
\end{equation}
where
\begin{equation}
\sigma_{l}=\sqrt{\sum\limits_{x=u-\varrho}^{x=u+\varrho}\sum\limits_{y=v-\varrho}^{y=v+\varrho}{i_{l}}^2(x,y)/n-{\mu_{l}}^2}
\label{eq.sigma_l_1}
\end{equation}
\begin{equation}
\sigma_{r}=\sqrt{\sum\limits_{x=u-\varrho}^{x=u+\varrho}\sum\limits_{y=v-\varrho}^{y=v+\varrho}{i_{r}}^2(x-d,y)/n-{\mu_{r}}^2}
\label{eq.sigma_r_1}
\end{equation}


$c(u,v,d)\in[-1,1]$ is the correlation cost. $i_l(x,y)$ denotes the intensity of a pixel at $(x,y)$ in the left image and $i_r(x-d,y)$ represents the intensity of a pixel at $(x-d,y)$ in the right image. The centre of the left block is $(u,v)$. Its width is $2\varrho+1$ and the number of pixels within each block is $n=(2\varrho+1)^2$. $\mu_l$ and $\mu_r$ represent the means of the pixel intensities within the left and right blocks, respectively. $\sigma_l$ and $\sigma_r$ denote the standard deviations of the left and right blocks, respectively. The calculations of $\mu$ and $\sigma$ can be accelerated using four integral images \cite{Evans2018}.


In a practical implementation, the values of $\mu_l$, $\mu_r$, $\sigma_l$ and $\sigma_r$ are pre-calculated and stored in static program storage for direct indexing \cite{Fan2018}. Therefore, only the dot product $\sum i_l i_r$ in Eq. (\ref{eq.ncc}) {needs to be calculated when computing the correlation cost between each pair of left and right blocks}. This greatly reduces the unnecessary computations.  Each computed correlation cost $c$ is stored in two 3-D cost volumes. It is to be noted that the value of $c$ at the position of $(u,v,d)$ in the reference cost volume is the same as that at the position of $(u-d,v,d)$ in the target cost volume. More details on the implementation of NCC are provided in \cite{Fan2017}.

\subsubsection{Cost Aggregation}
\label{sec.cost_aggregation}

In global algorithms, finding the best disparities is equivalent to {maximising} the joint probability in Eq. (\ref{eq.mrf_eq1}) \cite{Fan2018}:

\begin{equation}
P(\boldsymbol{p}, q)=\prod_{\boldsymbol{p_{ij}}\in\mathscr{P}} \Phi(\boldsymbol{p_{ij}}, q_{\boldsymbol{p_{ij}}})\prod_{\boldsymbol{n_{p_{ij}}}\in\mathscr{N}_{\boldsymbol{p_{ij}}}} \Psi (\boldsymbol{p_{ij}}, \boldsymbol{n_{p_{ij}}})
\label{eq.mrf_eq1}
\end{equation}
where $\boldsymbol{p_{ij}}$ represents a vertex at the site $(i,j)$ in the disparity map $\mathscr{P}$ and $q_{\boldsymbol{p_{ij}}}$ denotes the intensity differences which correspond to different disparities $d$. $\mathscr{N}_{\boldsymbol{p_{ij}}}=\{n_{1\boldsymbol{p_{ij}}},n_{2\boldsymbol{p_{ij}}},n_{3\boldsymbol{p_{ij}}},\cdots,n_{k\boldsymbol{p_{ij}}}|n_{\boldsymbol{p_{ij}}}\in\mathscr{P}\}$ represents the neighbourhood system {of} $\boldsymbol{p_{ij}}$. In this paper, the value of $k$ is set to $8$ and $\mathscr{N}$ is an 8-connected neighbourhood. $\Phi(\cdot)$ expresses the compatibility between  each possible disparity $d$ and the corresponding block similarity. $\Psi(\cdot)$ expresses the compatibility between $\boldsymbol{p_{ij}}$ and its neighbourhood system $\mathscr{N}_{\boldsymbol{p}_{ij}}$. However, Eq. (\ref{eq.mrf_eq1}) is commonly formulated  as an energy function, as follows \cite{Fan2018}:

\begin{equation}
\begin{split}
E(\boldsymbol{p})&=\sum_{\boldsymbol{p_{ij}}\in\mathscr{P}} D(\boldsymbol{p_{ij}}, q_{\boldsymbol{p_{ij}}})+
\sum_{\boldsymbol{n_{p_{ij}}}\in\mathscr{N}_{\boldsymbol{p_{ij}}}} V (\boldsymbol{p_{ij}}, \boldsymbol{n_{p_{ij}}})\\
\end{split}
\label{eq.mrf_eq2}   
\end{equation}
where $D(\cdot)$ corresponds to the correlation cost $c$ in this paper and $V(\cdot)$ determines the aggregation strategy. In the MRF model, the method to formulate an adaptive $V(\cdot)$ is important because the intensity in a discontinuous area usually greatly differs from {those} of its neighbours \cite{Li2012}. Therefore, some authors formulated $V(\cdot)$ as a piece-wise model to discriminate the discontinuous areas \cite{Kolmogorov2001}. However, the process of minimising the energy function in Eq. (\ref{eq.mrf_eq2}) results in a high computational complexity, making real-time performance challenging.
Since Tomasi et al. introduced the bilateral filter in \cite{Tomasi1998}, many authors have investigated its application to aggregate the matching costs \cite{ Yang2009, Hosni2013}. Mozerov and Weijer also proved that the bilateral filtering is a feasible solution for the energy minimisation problem in the MRF model \cite{Mozerov2015}. These methods are also known as fast bilateral stereo, where both  intensity difference and spatial distance provide a Gaussian weighting function to adaptively constrain the cost aggregation from the neighbours. A general representation of the cost aggregation in FBS is as follows \cite{Fan2018b}:
\begin{figure*}[!t]
	\begin{center}
		\centering
		\includegraphics[width=0.97\textwidth]{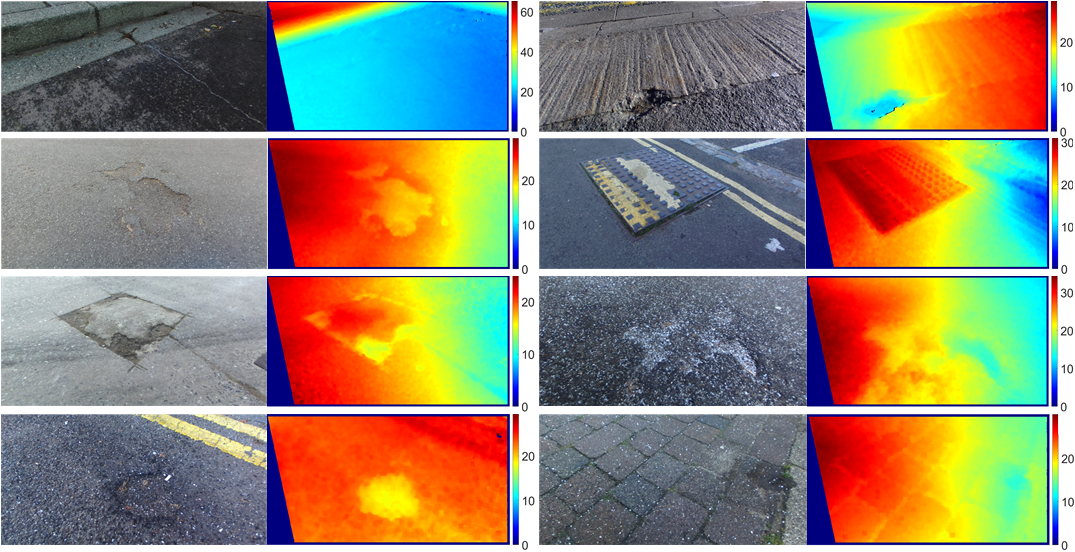}
		\caption{Experimental results. The first and third columns are the reference images. The second and forth columns are the subpixel disparity maps without post-processing.}
		\label{fig.exp_results}
	\end{center}
\end{figure*}

\begin{equation}
c_{agg}(u,v,d)=\frac{\sum_{x=u-\rho}^{x=u+\rho}  \sum_{y=v-\rho}^{y=v+\rho}  \omega_d(x,y)\omega_r(x,y)c(x,y,d)}{\sum_{x=u-\rho}^{x=u+\rho}  \sum_{y=v-\rho}^{y=v+\rho} \omega_d(x,y)\omega_r(x,y)}
\label{eq.fbs}
\end{equation}
where 
\begin{equation}
\omega_d(x,y)=\exp \bigg\{ -\frac{(x-u)^2+(y-v)^2}{{\gamma_d}^2} \bigg\}
\label{eq.omega_d}
\end{equation}
\begin{equation}
\omega_r(x,y)=\exp \bigg\{ -\frac{(i(x,y)-i(u,v))^2}{{\gamma_r}^2} \bigg\}
\label{eq.omega_r}
\end{equation}

$\omega_d$ and $\omega_r$ are based on the spatial distance and the colour similarity, respectively. $\gamma_d$ and $\gamma_r$ are two parameters used to control the values of $\omega_d$ and $\omega_r$, respectively. The costs $c$ within a square block are aggregated adaptively and an updated cost $c_{agg}$ can be obtained.

Although FBS has shown a good performance in terms of matching accuracy, it usually takes a long time to process the whole cost volume. Therefore, FBS must be implemented on powerful hardware in order to perform in real time. In this paper,  FBS is implemented on an NVIDIA GTX 1080 GPU. The performance of the implementation will be discussed in Section \ref{sec.experimental_results}. 

\subsubsection{Disparity Optimisation}

As discussed in Section \ref{sec.cost_aggregation}, the process of energy minimisation in global algorithms can be realised by performing bilateral filtering on the initial cost volumes. The best disparities can therefore be found by simply performing WTA on the left and right aggregated cost volumes. The reference and target disparity maps, i.e., $\ell^{ref}$ and $\ell^{tar}$, are then utilised to remove the incorrect matches.

\subsubsection{Disparity Refinement}
According to the uniqueness constraint stated in \cite{Fan2017}, for an arbitrary point $\boldsymbol{p_l}=[u_l,v_l]^\top$ in the reference image, there exists at most one corresponding point $\boldsymbol{p_r}=[u_r,v_r]^\top$ in the target image, namely:
\begin{equation}
\ell^{ref}(u,v)=\ell^{tar}(u-\ell^{ref}(u,v),v)
\end{equation}

Therefore, we first check the consistency of the left and right disparity maps to remove the occlusion areas \cite{Fan2017}.

Furthermore, since the value of $z^\mathscr{W}$ is inversely proportional to $d$, a disparity error larger than one pixel may result in a non-negligible difference in the 3-D reconstruction result \cite{Llorca2010}. Therefore, a subpixel enhancement is always performed to increase the resolution of the disparity maps. This can be achieved by fitting a parabola $f(u,v,d)$ to three correlation costs $c(u,v,d-1)$, $c(u,v,d)$ and $c(u,v,d+1)$ around the initial disparity $d$ and then selecting the centre line of $f(u,v,d)$ as the subpixel disparity $d^s$ \cite{Fan2018}, as shown in Eq. (\ref{eq.subpixel}). The corresponding subpixel disparity map is shown in Fig. \ref{fig.block_diagram}.

\begin{equation}
d^s=d+\frac{c(u,v,d-1)-c(u,v,d+1)}{2c(u,v,d-1)+2c(u,v,d+1)-4c(u,v,d)}
\label{eq.subpixel}
\end{equation}

\subsection{Post-Processing}
\label{sec.post-processing}

Due to the fact that the perspective views have been transformed in Section \ref{sec.perspective_transformation}, the estimated subpixel disparities on row $v$ should be added $\alpha_0+\alpha_1 v-\delta$ to obtain the post-processed disparity map (see Fig. \ref{fig.block_diagram}). The latter is then used to reconstruct the 3-D road surface.

\subsection{3-D Reconstruction}
\label{sec.3d_reconstruction}

Now, each 3-D point $\boldsymbol{p^\mathscr{W}}=[x^\mathscr{W},y^\mathscr{W},z^\mathscr{W}]^\top$ can be computed from its projections $\boldsymbol{{p}_l}=[u_l,v_l]^\top$ and $\boldsymbol{{p}_r}=[u_r,v_r]^\top$ using the intrinsic and extrinsic parameters of the stereo system, where $v_r$ is equivalent to $v_l$, and $u_r$ is associated with $u_l$ by $d$. The reconstructed 3-D road surface is as shown in Fig. \ref{fig.block_diagram}.

\section{Experimental Results}
\label{sec.experimental_results}

In this section, we evaluate the performance of the proposed road surface 3-D reconstruction algorithm both qualitatively and quantitatively. The algorithm is implemented on an NVIDIA GTX 1080 GPU for the purpose of real-time. We use the datasets described in our previously published work \cite{Fan2018} to evaluate the accuracy of the reconstructed 3-D road surface. Some examples of the disparity maps are shown in Fig. \ref{fig.exp_results}.

\begin{figure}[t!]
	\centering
	\subfigure[]
	{
		\includegraphics[width=0.225\textwidth]{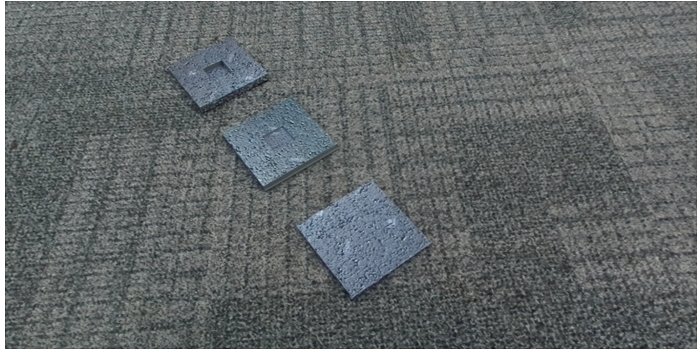}
	}
	\subfigure[]
	{
		\includegraphics[width=0.225\textwidth]{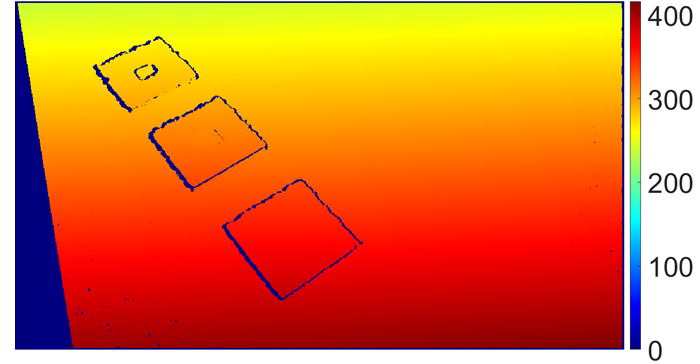}
	}
	\subfigure[]
{
	\includegraphics[width=0.25\textwidth]{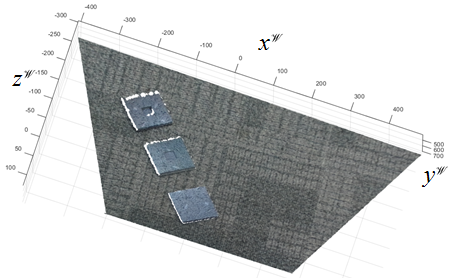}
}
\subfigure[]
{
	\includegraphics[width=0.20\textwidth]{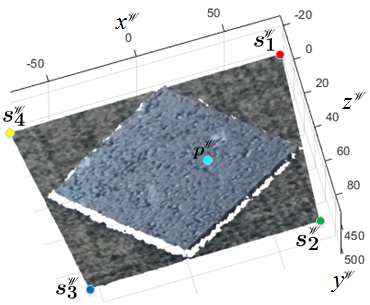}
}
	\caption{Sample model 3-D reconstruction. (a) reference image. (b) subpixel disparity map with post-processing. (c) reconstructed scenery. (d) selected 3-D point cloud which includes one of the sample models.}
	\label{fig.reconstruction_eva}
	\vspace{-1.9em}
\end{figure}

To quantify the accuracy of the reconstructed road surface, we designed some sample models with different sizes which were printed using a MakerBot Replicator 2 Desktop 3-D Printer. More details on these sample models are provided in \cite{Fan2018}. An example of the reference image which contains the sample models is illustrated in Fig. \ref{fig.reconstruction_eva} (a). The corresponding post-processed disparity map and 3-D reconstruction result are shown in Fig. \ref{fig.reconstruction_eva} (b) and (c), respectively. To evaluate the reconstruction accuracy,  we randomly select a region which includes one of the sample models from Fig. \ref{fig.reconstruction_eva} (c) (see Fig. \ref{fig.reconstruction_eva} (d)). The four corners $\boldsymbol{s_1^\mathscr{W}}$, $\boldsymbol{s_2^\mathscr{W}}$, $\boldsymbol{s_3^\mathscr{W}}$ and $\boldsymbol{s_4^\mathscr{W}}$ are interpolated into a ground plane $n_0x^\mathscr{W}+n_1y^\mathscr{W}+n_2z^\mathscr{W}+n_3=0$ which is considered as the road surface. Then, we select some random 3-D points $\boldsymbol{p_1^\mathscr{W}}$, $\boldsymbol{p_2^\mathscr{W}}$, ..., $\boldsymbol{p_n^\mathscr{W}}$ on the surface of the model and estimate the distances between them and the road surface. These random distances provide a measurement range of the model height. Based on our experimental results, the reconstruction accuracy is approximately 3 mm. Since the baseline of the ZED camera is fixed and cannot be increased to improve the reconstruction accuracy, we mount the stereo rig to a relatively low height and keep it as perpendicular as possible to the road surface to reduce the average depth. This ensures a high reconstruction accuracy. 

Furthermore, the execution speed of the proposed algorithm is also quantified in order to give an evaluation of the overall system's performance. However, due to the fact that image size and disparity range are not constant for different datasets, a general way to depict the performance in terms of processing speed is given in millions of disparity evaluations per second $Mde/s$ as follows \cite{Tippetts2016}: 

\begin{equation}
Mde/s=\frac{u_{max}v_{max}d_{max}}{t}{10^{-6}}
\label{eq.mde_s}
\end{equation}
where $u_{max}$ and $v_{max}$ represent the width and height of the disparity map, $d_{max}$ denotes the maximum search range and $t$ represents the algorithm runtime in seconds. Our implementation achieves a performance of $566.37$ Mde/s when $\varrho$ and $\rho$ are set to $3$ and $4$, respectively (image resolution: $1240\times609$).

\begin{figure}[!t]
	\begin{center}
		\centering
		\includegraphics[width=0.34\textwidth]{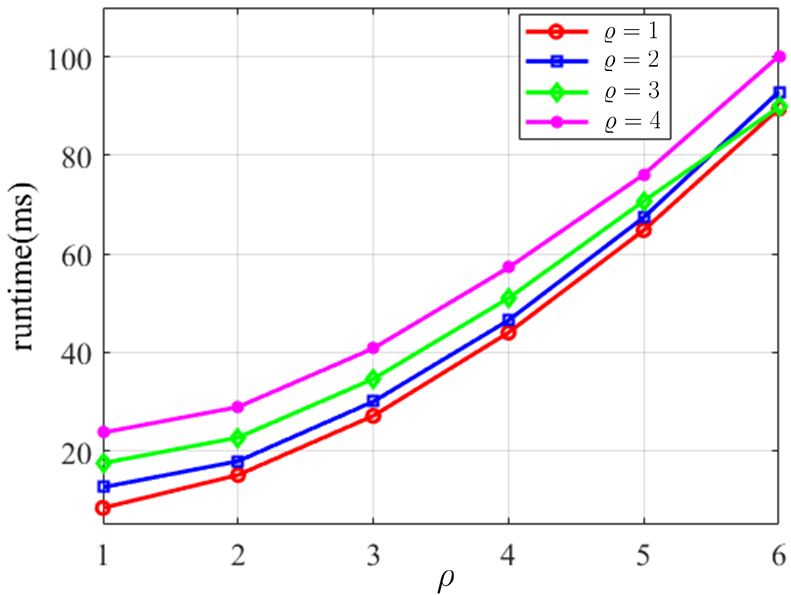}
		\caption{Runtime with respect to different values of $\varrho$ and $\rho$.}
		\label{fig.eva4}
	\end{center}
\end{figure}

Moreover, we evaluate the speed performance with respect to different values of $\varrho$ and $\rho$.
The runtime $t$ corresponding to different values of $\varrho$ and $\rho$ is shown in Fig. \ref{fig.eva4}. It can be seen that $t$ increases with increasing $\varrho$ or $\rho$. When $\rho$ is equal to $2$, the execution speed of the proposed implementation is around $50$ fps. When $\rho$ is set to $4$, the processing time is doubled (see Fig. \ref{fig.eva4}). To achieve a better trade-off between accuracy and speed, we set $\varrho$ and $\rho$ to $3$ and $4$, respectively. 

\section{Conclusion and Future Work}
\label{sec.conclusion_future_work}

In this paper, we present a real-time stereo vision system for road surface 3-D reconstruction. By performing the perspective transformation before disparity estimation, the road scenery in the reference and target images become more similar. Compared with our previously published disparity estimation algorithm \cite{Fan2018}, FBS is more capable of running in parallel. Therefore, the proposed road surface 3-D reconstruction algorithm is implemented on an NVIDIA GTX 1080 GPU for the real-time purpose. According to the experimental results, the implementation on the GPU achieves a processing speed of 25 fps and the reconstruction accuracy is approximately 3 mm.  

However, performing the bilateral filtering on the whole cost volume is a very time-consuming process. Therefore, we plan to reduce the search range and only perform the bilateral filtering on the space around the calculated disparities. Furthermore, the reconstructed road sceneries will be used for road Simultaneous Localisation and Mapping (SLAM). Finally, we plan to transform the disparity map using the algorithm presented in \cite{Fan2018c} to detect potholes. 

\bibliographystyle{IEEEbib}

\end{document}